\documentclass[lettersize,journal]{IEEEtran}
\usepackage{amsmath,amsfonts}
\usepackage{algorithmic}
\usepackage{algorithm}
\usepackage{array}
\usepackage[caption=false,font=normalsize,labelfont=sf,textfont=sf]{subfig}
\usepackage{textcomp}
\usepackage{stfloats}
\usepackage{url}
\usepackage{verbatim}
\usepackage{graphicx}
\usepackage{multirow}
\usepackage{booktabs}
\begin{document}

\title{Attack for Defense: Adversarial Agents for Point Prompt Optimization Empowering Segment Anything Model}

\author{Xueyu Liu,~\IEEEmembership{~IEEE Member}, Xiaoyi Zhang, Guangze Shi, Meilin Liu, Yexin Lai, Yongfei Wu,~\IEEEmembership{~IEEE Member}, Mingqiang Wei,~\IEEEmembership{~IEEE Senior Member}
		\thanks{This work was supported by the National Natural Science Foundation of China (Grant Nos. 11472184, 11771321, 61901292); the National Youth Science Foundation of China (Grant No.11401423); the ShanXi province plan project on Science and Technology of social Development (Grant No. 201703D321032);  and the Natural Science Foundation of Shanxi Province, China (Grant No. 201901D211080).  (Xueyu Liu and Xiaoyi Zhang contributed equally to this work). (Corresponding authors: Yongfei Wu).}
\thanks{Xueyu Liu,  Xiaoyi Zhang, Guangze Shi, Meilin Liu, Yexin Lai, Yongfei Wu are with the College of Artificial Intelligence, Taiyuan University of Technology, Taiyuan, 030024, China (e-mail: wuyongfei@tyut.edu.cn). }
\thanks{Mingqiang Wei is with the School of Data Science, Nanjing University Of Aeronautics And Astronautics, Nanjing, 210016, China (e-mail: mqwei@nuaa.edu.cn).}}

\markboth{Journal of \LaTeX\ Class Files,~Vol.~14, No.~8, August~2021}%
{Shell \MakeLowercase{\textit{et al.}}: A Sample Article Using IEEEtran.cls for IEEE Journals}


\maketitle

\begin{abstract}
Prompt quality plays a critical role in the performance of the Segment Anything Model (SAM), yet existing approaches often rely on heuristic or manually crafted prompts, limiting scalability and generalization. In this paper, we propose Point Prompt Defender, an adversarial reinforcement learning framework that adopts an attack-for-defense paradigm to automatically optimize point prompts. We construct a task-agnostic point prompt environment by representing image patches as nodes in a dual-space graph, where edges encode both physical and semantic distances. Within this environment, an attacker agent learns to activate a subset of prompts that maximally degrade SAM’s segmentation performance, while a defender agent learns to suppress these disruptive prompts and restore accuracy. Both agents are trained using Deep Q-Networks with a reward signal based on segmentation quality variation. During inference, only the defender is deployed to refine arbitrary coarse prompt sets, enabling enhanced SAM segmentation performance across diverse tasks without retraining. Extensive experiments show that Point Prompt Defender effectively improves SAM’s robustness and generalization, establishing a flexible, interpretable, and plug-and-play framework for prompt-based segmentation.
\end{abstract}


\section{Introduction}

In recent years, the dominant paradigm for image segmentation has relied on architecture engineering—designing specialized network structures such as U-Net \cite{ronneberger2015u} and training them on task-specific datasets. While this paradigm has led to significant performance gains, it faces critical limitations: poor generalization to unseen domains, susceptibility to local optima, limited task transferability, and high costs in architecture design and tuning. These challenges are especially pronounced in scenarios with scarce labels, shifting data distributions, or time-sensitive applications.

Recent advances in pretrained foundation models (PFMs) have significantly expanded the capabilities of computer vision systems, particularly in achieving domain-robust segmentation across diverse scenarios \cite{radford2021learning, jia2021scaling}. Vision foundation models (VFMs), such as DINOv2 \cite{oquab2023dinov2, caron2021emerging}, learn generic and transferable visual representations by capturing rich information at both the image and pixel levels, solely from raw image data. This enables strong generalization and zero-shot performance across a wide range of downstream tasks. Among these, the Segment Anything Model (SAM) \cite{kirillov2023segment, ravi2024sam} has emerged as a leading framework for general-purpose segmentation. Through large-scale pretraining on diverse mask–image–prompt triplets and a prompt-driven interaction design, SAM decouples inference from task-specific supervision, offering excellent adaptability and training-free deployment. Despite its strengths, SAM still depends heavily on manual prompt input to achieve accurate segmentation. This reliance on human interaction limits its scalability in fully automated pipelines or in applications where user input is impractical or costly.


\begin{figure}[t]
	\centering
	\includegraphics[width=0.95\columnwidth]{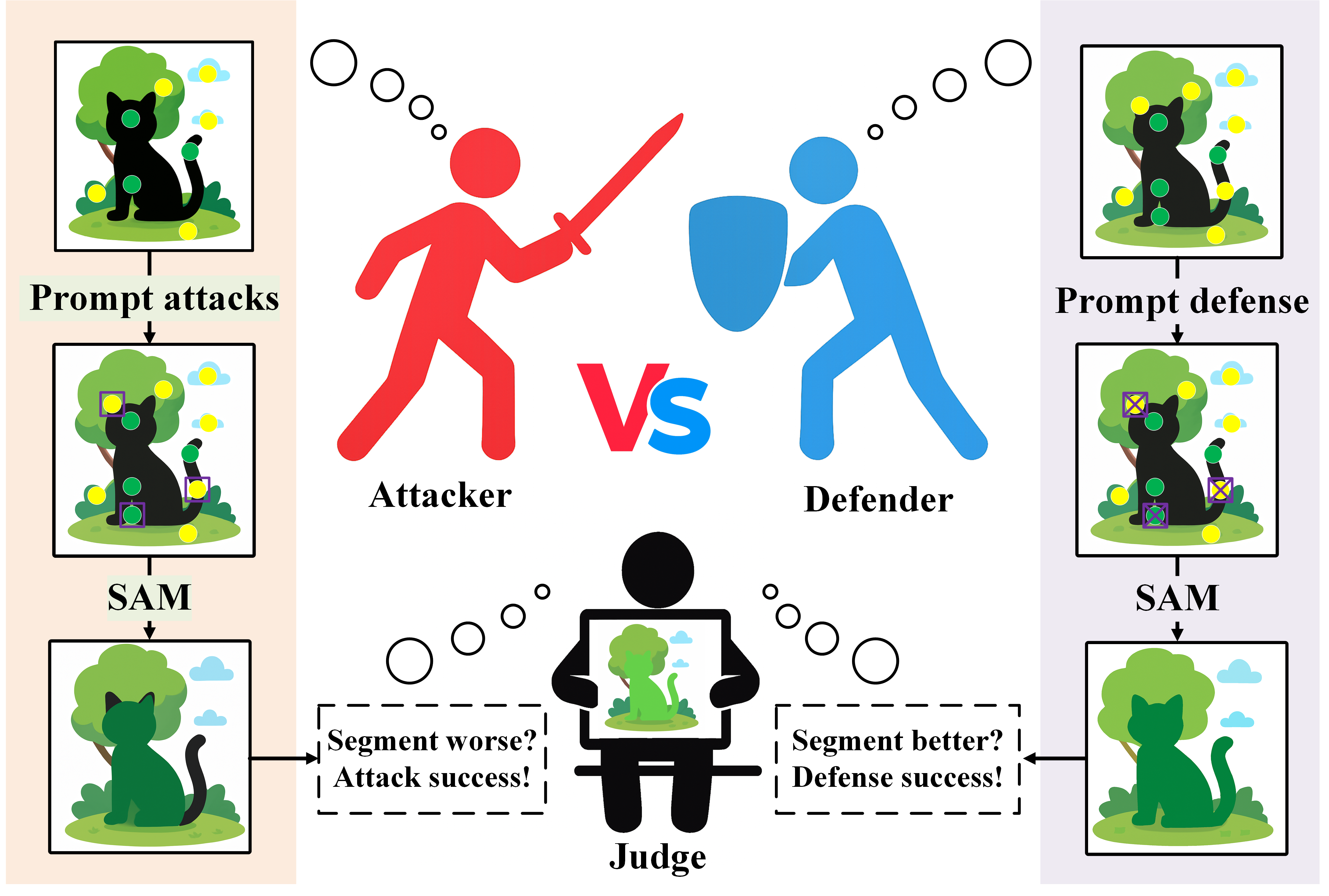} 
	\caption{Adversarial training in PPD: The attack agent activates prompts to worsen SAM segmentation, while the defense agent deactivates harmful prompts to improve it. A judge evaluates the segmentation quality based on the ground truth.}
	\label{fig_eg}
\end{figure}

Several recent efforts have explored automatic or learned prompt generation techniques to enhance SAM’s usability. One line of work focuses on fine-tuning prompt encoders to adapt SAM to downstream segmentation tasks \cite{zhang2024personalize, sun2024vrp, huang2024alignsam, kweon2024sam, peng2024sam, kato2024generalized, li2025stitching}, but such approaches often require additional task-specific supervision, hindering scalability to unseen domains. To overcome this, geometric prompt engineering methods—such as Matcher \cite{liu2024matcher}, GBMSeg \cite{liu2024featureprompting}, and others \cite{wang2025osam, liu2025segment}—have been proposed to generate prompts heuristically based on spatial or semantic priors. While these methods improve automation, they treat prompts as static inputs, lacking the ability to adaptively reason about prompt utility across varying segmentation contexts. To address the limitations of static prompt strategies, Point Prompt Optimizer (PPO) \cite{liu2025plug} introduces a reinforcement learning framework to iteratively refine prompt locations based on inter-prompt relations. However, PPO optimizes prompts in isolation from SAM’s segmentation outputs, without directly coordinating with SAM’s feedback during training. Moreover, due to the limited diversity of its training environments, the learned policies are highly sensitive to the quality and distribution of initial prompts, often failing to generalize across domains or scene structures.

In this paper, we propose Point Prompt Defender (PPD), a novel adversarial reinforcement learning framework for optimizing point prompts in SAM. As shown in Figure \ref{fig_eg}, unlike prior methods that passively generate or rank prompts, PPD actively explores the segmentation landscape through a competitive two-agent setup: an attack agent learns to activate prompts that degrade SAM’s segmentation performance, while a defense agent learns to suppress disruptive prompts and recover accuracy. This interaction is modeled as a multi-agent game within a task-agnostic dual-space graph environment constructed using features extracted by DINOv2. In this graph, image patches are represented as nodes, and edges encode both feature similarity and physical proximity. The agents operate over both explicit prompt actions (e.g., activation and deactivation) and implicit structural cues (e.g., feature and physical distances), enabling joint reasoning over prompt configurations and spatial topology. Both agents are trained using Deep Q-Networks (DQN), with reward signals driven by variations in segmentation quality.

At inference time, only the defense agent is deployed to refine arbitrary prompt sets in a task-agnostic and fully automatic manner. By coupling prompt optimization directly with SAM’s segmentation feedback, PPD departs from static or supervised strategies and learns to suppress adversarial prompts in a dynamic environment induced by the attacker. This leads to strong generalization across varying prompt configurations and downstream tasks, without requiring retraining. Our main contributions are summarized as follows:

\begin{itemize}
	
	
	

	\item We propose PPD, an adversarial reinforcement learning framework with an attack-for-defense strategy for optimizing point prompts in SAM.
	
	\item PPD leverages segmentation feedback to guide agents through explicit actions and implicit structural information for prompt-based attack and defense.
	
	\item PPD enables task-agnostic prompt optimization at inference by deploying only the defense agent, effectively suppressing harmful prompts without retraining.

\end{itemize}

\section{Related Work}
\subsection{Vision Foundation Models}

Recent advances in VFMs have led to remarkable improvements in various downstream tasks. Among them, DINOv2~\cite{oquab2023dinov2} stands out as a self-supervised framework that learns transferable visual representations without any labels, achieving competitive performance on both image-level and dense prediction tasks. Unlike supervised models, DINOv2 leverages large-scale curated data and knowledge distillation to construct scalable, general-purpose visual encoders. In parallel, promptable segmentation models have gained increasing attention. The most notable one is the SAM~\cite{kirillov2023segment}, which formulates segmentation as a prompt-based task and demonstrates strong generalization on diverse domains through zero-shot inference. SAM is trained with over 1 billion masks, marking a significant milestone in open-vocabulary segmentation. Recent works further analyze SAM's extensibility~\cite{zhang2023survey} and performance bottlenecks~\cite{zhang2023comprehensive}, while SAM2~\cite{ravi2024sam2} extends SAM to video segmentation with improved efficiency and memory mechanisms. Although these VFMs provide powerful feature representations and flexible interfaces for segmentation, their reliance on hand-crafted prompts or simple heuristics limits performance in few-shot and noisy scenarios, motivating our study on learning to automatically optimize prompts via reinforcement learning.

\subsection{Vision Prompt Engineering}
In natural language processing (NLP), a wide range of studies have explored automatic prompt design techniques to improve the effectiveness of PFMs. These include various prompt tuning frameworks that learn optimal textual prompts to guide model behavior more effectively across downstream tasks \cite{zhou2022conditional, jia2022visual, bahng2022exploring, khattak2023maple, yao2023visual, lei2024prompt, qiu2024progressive}. Inspired by this success, the vision community has also turned its attention to prompt engineering strategies, particularly in the context of adapting the SAM for diverse segmentation scenarios.

One prominent line of work focuses on training learnable prompt encoders, which embed high-level visual cues or task-specific semantics into prompt representations for SAM's decoder \cite{shaharabany2023autosam, xie2024masksam, huang2024learning, sun2024vrp}. For instance, VRP-SAM introduces a visual reference prompt encoder to translate various reference signals into meaningful prompt embeddings, enabling more context-aware mask generation \cite{sun2024vrp}. Although effective, these methods are often limited by their dependency on downstream task supervision or task-specific finetuning, which reduces their ability to generalize across domains or tasks.

An alternative direction involves directly generating geometric prompts (e.g., point or box prompts) to control SAM’s segmentation output, which can be roughly categorized into two strategies. The first strategy leverages a small amount of annotated data to train auxiliary segmentation models, from which prompts are sampled based on the generated pseudo-labels \cite{wang2023review, na2024segment, zhang2023survey, zhang2024personalize, chen2024sam}. Li et al., for example, used coarse predictions to randomly sample points as input prompts to SAM after fine-tuning \cite{li2023auto}, while Wang et al. adopted a prototype-based learning framework to extract high-confidence prompts \cite{wang2023mathrm}. Dai et al. further enriched this line by exploring single-point prompt augmentation strategies to enhance prompt diversity and utility \cite{dai2023samaug}. However, these approaches still rely on task-specific training, limiting their scalability to unseen domains. However, these methods share two key limitations: they ignore SAM’s segmentation feedback during optimization, and their performance is highly sensitive to the quality and layout of initial prompts, limiting generalization across domains.

To address these issues, PPD introduces a dual-agent system, where an attack agent deliberately applies diverse disruptive prompts to simulate challenging conditions, and a defense agent learns to counteract them by refining prompts to improve segmentation outcomes. The system is trained using the change in DICE score as a reward signal, allowing both agents to adaptively optimize prompt configurations based on segmentation feedback in the training phase. This adversarial setup strengthens the robustness and generalization of prompt refinement without requiring task-specific training in the inference phase.

\begin{figure*}[t]
	\centering
	\includegraphics[width=1.9\columnwidth]{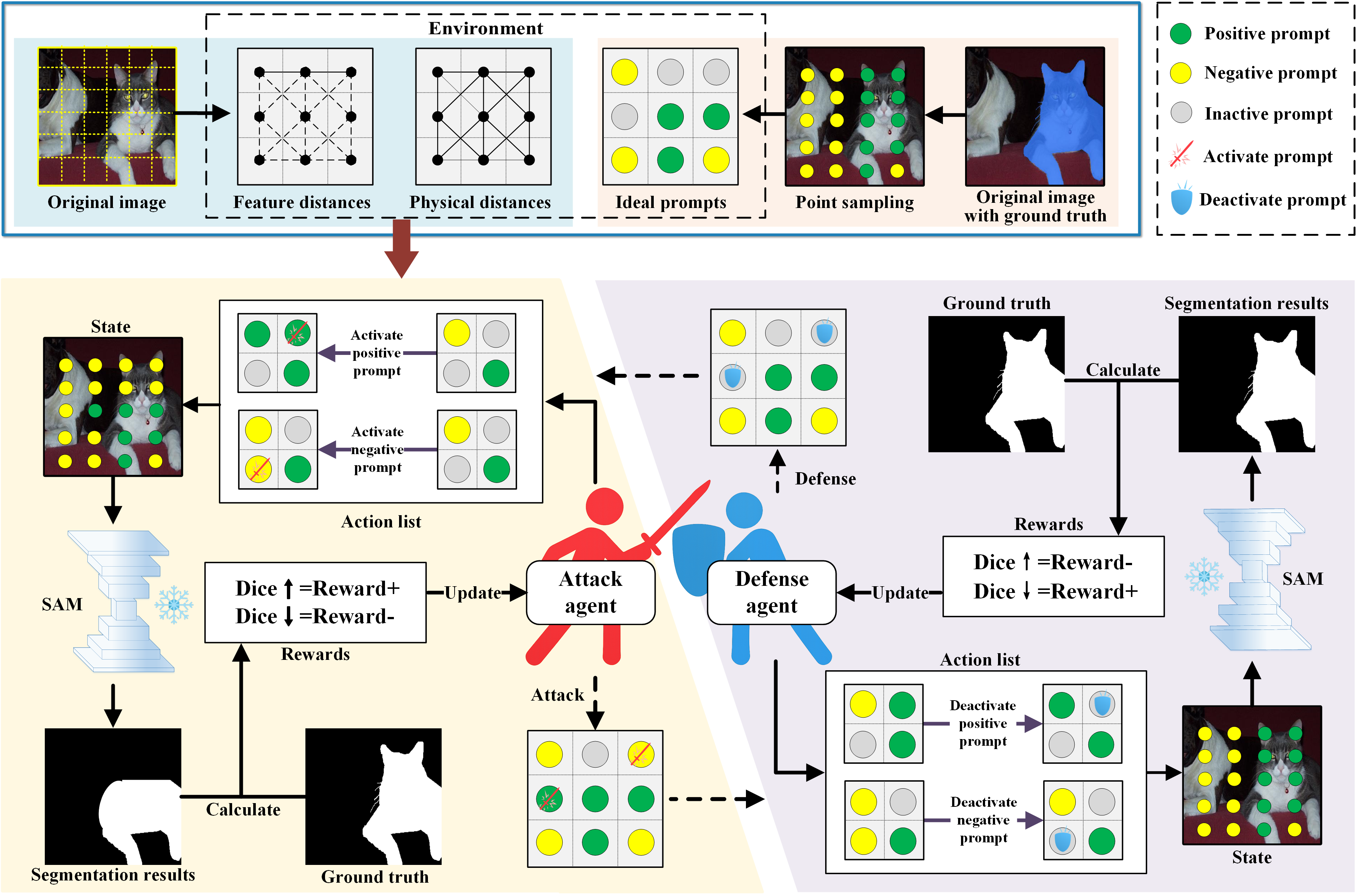} 
	\caption{Overview of PPD: A dual-space graph and ideal prompts form the environment. Guided by SAM segmentation feedback, the attack agent activates poor prompts to degrade performance, while the defense agent suppresses them to recover accuracy. Solid and dashed lines denote agent training and testing phases, respectively.}
	\label{fig_structure}
\end{figure*}
\section{Method}

This section introduces PPD, an adversarial reinforcement learning framework that adopts an attack-for-defense paradigm to optimize point prompts for SAM. An overview of PPD is illustrated in Figure~\ref{fig_structure}. Specifically, we construct a task-agnostic prompt optimization environment by representing image patches as nodes in a dual-space graph, where edges encode both feature and physical distances. Ideal Prompt Initialization is used to initialize this environment by generating prompt points based on the ground-truth segmentation mask. Positive prompts are uniformly sampled within the mask, while negative prompts are sampled outside the mask. Within this environment, an attack agent learns to activate a subset of prompts that maximally degrade SAM’s segmentation performance, while a defense agent learns to suppress these disruptive prompts and restore accuracy. Both agents are trained using Deep Q-Networks with reward signals derived from segmentation quality variation. During inference, only the defense agent is retained to refine arbitrary coarse prompt sets, enabling enhanced SAM segmentation across diverse tasks without retraining. In the following subsections, we describe each component in detail.

\subsection{Environment Construction}

To support structured prompt optimization via reinforcement learning, we define the environment as a dual-space heterogeneous graph equipped with both feature and physical edge attributes. The environment is initialized using ideal prompts derived from ground-truth segmentation masks. This prompt-aware environment seamlessly integrates explicit prompt operations with implicit structural relations among prompts, enabling both direct manipulation and relational reasoning for effective policy learning.

\paragraph{Dual-space distance matrix calculation.} Given an input image $X$, we divide it into a set of non-overlapping or sliding patches $x = \{x_1, x_2, \ldots, x_n\}$. Each patch $x_i$ is represented by a feature vector $f_i$ extracted using the DINOv2 image encoder~\cite{oquab2023dinov2}, resulting in $f = \{f_1, f_2, \ldots, f_n\}$. Based on these features, we construct a feature distance matrix $M_f$ by computing pairwise Euclidean distances between patch embeddings:
\begin{equation}
	M_f(i, j) = \left\lVert f_i - f_j \right\rVert.
\end{equation}
Simultaneously, we compute a physical distance matrix $M_p$ by measuring the Euclidean distances between the geometric centers of patches:
\begin{equation}
	M_p(i, j) = \left\lVert x_i - x_j \right\rVert.
\end{equation}
These two matrices jointly encode semantic and spatial affinities and are used to define the edge attributes in a dual-space graph $G = (V, E_f, E_p)$, where $E_f$ and $E_p$ represent feature-based and physical-based edge sets, respectively.

\paragraph{Ideal prompt initialization.} To initialize the environment with a meaningful prompt configuration, we generate an ideal prompt scheme from the ground-truth segmentation mask of each training image. Specifically, we uniformly sample points within and outside the mask region at fixed intervals. Points sampled inside the segmentation mask are assigned as positive prompts, while those outside the mask are treated as negative prompts. This ideal prompt distribution provides a high-quality initialization that guides the agent's adversarial learning. By using these prompts as the initial node set in the graph, the environment starts from a semantically informative state, enabling the attack and defense agents to focus on modifying the most impactful prompts for improving or degrading SAM's segmentation output.

\subsection{Adversarial Agent Training Process}

The primary goal of the PPD is to optimize the spatial distribution of point prompts by employing a two-stage adversarial interaction between an attack agent and a defense agent. In each training episode, the environment is initialized using the ideal prompt configuration, derived from the ground-truth segmentation mask. The attack agent begins by activating a subset of both positive and negative prompts from the inactive prompt pool, intentionally degrading SAM’s segmentation performance. This altered prompt set is then passed to SAM to generate a segmentation prediction. Following the attack, the defense agent takes over and receives the modified prompt configuration. The defense agent's task is to deactivate a selected set of active prompts, aiming to improve the segmentation quality. This refined set of prompts is inputted into SAM for a new segmentation prediction. The primary objective of the defense agent is to recover SAM's performance by deactivating harmful prompts introduced by the attacker. This interaction process is visualized in the Algorithm \ref{alg:ppd}.

\begin{algorithm}[ht]
	\caption{Training Procedure of PPD}
	\label{alg:ppd}
	\textbf{Input}: Dual-space graph $G$, ideal prompts $P_i$ and ground truth $M$ for each image $X$; SAM model \\
	\textbf{Parameters}: Max epochs $E$; Max steps per epoch $T$; \\
	\textbf{Output}: Trained $Q_{\text{att}}, Q_{\text{def}}$
	\begin{algorithmic}[1]
		\FOR{epoch = 1 to $E$}
		\STATE Select a set of $X$, $M$, $G$ and $P_i$ as the environment.
		\STATE \textbf{//Attack Phase:}
		\FOR{step = 1 to $T$}		
		\STATE Action list of attack agent=$\mathcal{A}_t^{\operatorname{\text{att}}}$
		\STATE Run SAM with new prompts to get $\hat{M}_t$
		\STATE Compute reward  $r_t^{\operatorname{\text{att}}} $ and update $Q_{\text{att}}$
		\ENDFOR
		\STATE Inference $Q_{\text{att}}$ on $P_i$ to get $P_{\text{att}}$ as the environment
		
		\STATE \textbf{//Defense Phase:}
		\FOR{step = 1 to $T$}
		\STATE Action list of defense agent=$\mathcal{A}_t^{\operatorname{\text{def}}}$
		\STATE  Run SAM with new prompts to get $\hat{M}_t $
		\STATE Compute reward $r_{\text{def}}$ and update $Q_{\text{def}}$
		
		\ENDFOR
		\ENDFOR
		\STATE \textbf{return} $Q_{\text{att}}, Q_{\text{def}}$
	\end{algorithmic}
\end{algorithm}

\paragraph{Attacker}  
The attack agent's goal is to degrade the segmentation performance of SAM by activating a set of point prompts. At each time step $t$, the attack agent selects an action $a_t^{\operatorname{\text{atk}}}$ from the space of inactive prompts:
\begin{equation}
	\mathcal{A}_t^{\operatorname{\text{atk}}} = \{ p_i \in \mathcal{P} \mid \text{status}_i = \text{inactive} \}.
\end{equation}
This action corresponds to the activation of several inactive prompts, which are then added to the current prompt configuration. The activated prompts are passed to SAM, which generates a segmentation mask $\hat{M}_t$. The attacker's objective is to disrupt SAM’s segmentation quality, and thus, the reward is designed to encourage performance degradation.

The reward function for the attack agent is based on the change in segmentation accuracy, as measured by the Dice coefficient. Specifically, the reward is defined as:
\begin{equation}
	r_t^{\operatorname{\text{atk}}} = -\left( \text{Dice}(\hat{M}_t, M) - \text{Dice}(\hat{M}_{t-1}, M) \right),
\end{equation}
where $\hat{M}_t$ is the segmentation mask obtained after the current attack, and $M$ is the ground truth segmentation mask. The Dice coefficient measures the overlap between the predicted segmentation and the ground truth. A decrease in the Dice coefficient (i.e., a worse segmentation result) yields a higher reward for the attacker. This incentivizes the attack agent to find the most effective points to activate, thereby maximizing the degradation of SAM's segmentation.

\paragraph{Defender}  
The defender's goal is to recover segmentation performance by deactivating harmful point prompts activated by the attacker. At each time step $t$, the defense agent selects an action $a_t^{\operatorname{\text{def}}}$ from the set of active prompts:
\begin{equation}
	\mathcal{A}_t^{\operatorname{\text{def}}} = \{ p_i \in \mathcal{P} \mid \text{status}_i = \text{active} \}.
\end{equation}
This action corresponds to the deactivation of a subset of previously activated prompts. The modified prompt set is then input into SAM to generate a new segmentation prediction $\hat{M}_t$. The defender’s objective is to improve SAM’s segmentation quality, and thus, the reward is designed to encourage performance recovery.

The reward function for the defense agent is based on the change in segmentation accuracy, as measured by the Dice coefficient. Specifically, the reward is defined as:
\begin{equation}
	r_t^{\operatorname{\text{def}}} = \text{Dice}(\hat{M}_t, M) - \text{Dice}(\hat{M}_{t-1}, M),
\end{equation}
where $\hat{M}_t$ is the segmentation mask obtained after the defender's action, and $M$ is the ground truth segmentation mask. The Dice coefficient measures the overlap between the predicted segmentation and the ground truth. An increase in the Dice coefficient (i.e., a better segmentation result) yields a higher reward for the defender. This incentivizes the defense agent to identify and deactivate harmful prompts that degrade segmentation quality, thereby improving SAM's performance.

\paragraph{Training process}

Both agents, the attack agent and the defense agent, are trained alternately in an adversarial fashion. In each training step, the attack agent is first updated by learning to identify and activate the most vulnerable prompts that degrade SAM's segmentation performance. After the attack agent's update, the defense agent is then trained to learn how to effectively remove the harmful prompts introduced by the attacker and restore segmentation accuracy.

The agents interact with a dual-space graph environment, and their actions are guided by DQN \cite{mnih2015human}. Each agent maintains a Q-network that approximates the action-value function, which maps states to expected rewards for each possible action. The Q-network for each agent is trained by minimizing the temporal difference loss:
\begin{equation}
	\mathcal{L}_t = \left( r_t + \gamma \max_{a'} Q_{\theta^-}(s_{t+1}, a') - Q_\theta(s_t, a_t) \right)^2,
\end{equation}
where \(r_t\) is the reward, \(s_t\) is the state at time step \(t\), and \(a_t\) is the action selected by the agent. The discount factor \(\gamma\) balances the importance of immediate and future rewards. The target network \(Q_{\theta^-}\) is periodically updated to stabilize training of $Q_{\text{att}}$ and $Q_{\text{def}}$.


Through this adversarial training paradigm, both agents converge towards an optimal solution, where the attack agent finds the most impactful prompts for degradation, and the defense agent identifies the best prompts for restoring segmentation performance. The use of DQN allows the agents to learn effective policies by interacting with the environment, where changes in the prompt set (i.e., activation or deactivation of points) lead to modifications in the structure of the graph $G$. These structural changes implicitly guide the agents to learn the most effective attack and defense strategies, thereby optimizing the prompt configuration in a task-agnostic manner.

\subsection{Agent inference for Prompt Optimization}

During inference, our method remains independent of any downstream task, as the environment is constructed entirely from task-agnostic features extracted by DINOv2~\cite{oquab2023dinov2}. Given any initial prompts, the defense agent filters out low-quality ones to enhance SAM's segmentation. This design enables a task-agnostic, plug-and-play enhancement without the need for additional retraining.

\begin{table*}[htpb]
	
	\centering
	\caption{Ablation results on natural and medical datasets. Top: degradation by attacks and recovery by defense. Bottom: performance gains from PPD optimization over feature matching.}
	\begin{tabular}{lllllll}
		\toprule
		\multirow{2}{*}{Methods} 
		& \multicolumn{2}{c}{PASCAL VOC}  
		& \multicolumn{2}{c}{ISIC}                                                     
		& \multicolumn{2}{c}{Kvasir}                                                                                                
		\\
		\cmidrule(lr){2-3} \cmidrule(lr){4-5} \cmidrule(lr){6-7}
		& \multicolumn{1}{c}{mDSC (\%$\uparrow$)} 
		& \multicolumn{1}{c}{mIoU (\%$\uparrow$)} 
		& \multicolumn{1}{c}{mDSC (\%$\uparrow$)} 
		& \multicolumn{1}{c}{mIoU (\%$\uparrow$)} 
		& \multicolumn{1}{c}{mDSC (\%$\uparrow$)} 
		& \multicolumn{1}{c}{mIoU (\%$\uparrow$)} 
		\\ 
		\midrule
		
		Ideal prompts &78.5&69.4&87.3&78.7&83.0&73.9 \\
		Attack ideal prompts &34.2 (-44.3) &21.5 (-47.9) &56.4 (-30.9) &48.3 (-30.4) &59.7 (-23.3) &48.9 (-25.0) \\
		Defense against attacks &73.5 (+39.3) &63.5 (+42.0) &81.5 (+25.1) &73.4 (+25.1) &76.4 (+16.7) &69.9 (+21.0) \\\midrule
		Feature matching         &41.3&34.8&66.4&55.0&31.4&21.5  \\
		Feature matching+PPD &69.1 (+27.8)&60.3 (+25.5)&76.3 (+9.9)&64.2 (+9.2)&54.8 (+23.4)&44.9 (+23.4) \\
		
		\bottomrule
	\end{tabular}

	\label{tab:attack_defense}
\end{table*}

\section{Experiments}
\subsection{Experiment Settings}
\paragraph{Datasets}We train the PPD framework using 1000 images randomly sampled from the FSS-1000 dataset \cite{li2020fss}, covering a wide range of object categories and appearances. This setup enables the model to learn class-agnostic prompt manipulation strategies under diverse visual conditions. After training, the learned PPD is directly applied—without any retraining or adaptation—to three benchmark datasets for evaluation: PASCAL VOC \cite{everingham2010pascal} for natural scenes, and ISIC \cite{shiraishi2000development} and Kvasir \cite{jha2020kvasir} for medical scenes. This training-free, one-shot evaluation highlights the plug-and-play capability and strong generalization performance of our method across multi-domain segmentation tasks.

\paragraph{Implementation details}
All experiments are conducted on Linux servers with 10 NVIDIA Tesla V100 GPUs. PPD is trained for 1000 episodes, each with a random number of environment steps sampled between 50 and 300 to simulate diverse initial prompt configurations. The Q-networks of both agents are optimized using Adam with a learning rate of $10^{-4}$ and batch size 128. Target networks are updated every 100 environment steps based on a global counter. An $\epsilon$-greedy strategy is used, with $\epsilon$ linearly annealed from 1.0 to 0.1 during training.

\subsection{Ablation Study}
To assess the task-agnostic capability of PPD, we evaluate it on three datasets excluded from the training phase.

\begin{figure}[htp]
	\centering

	\includegraphics[width=0.93\linewidth]{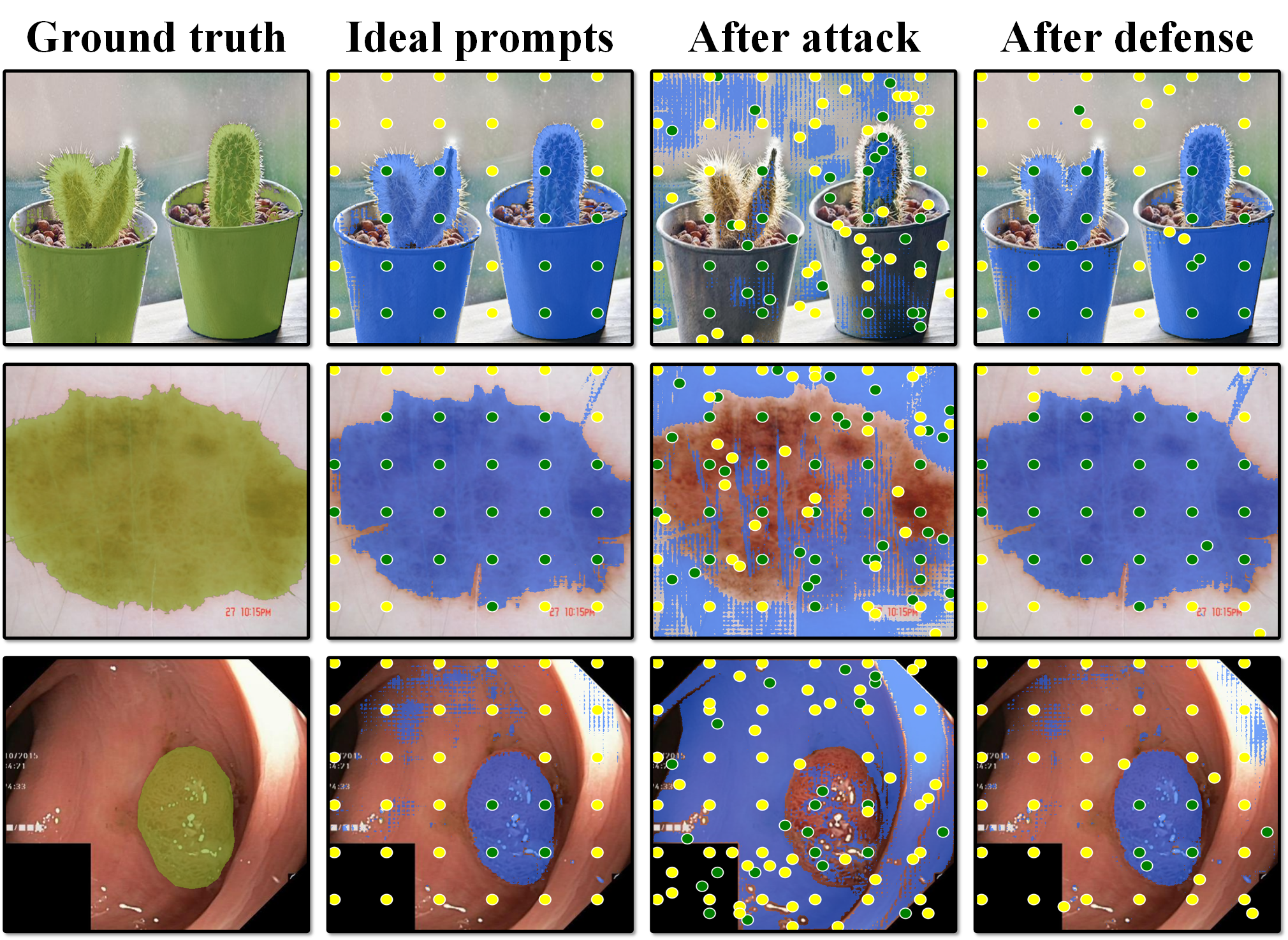} 
	\caption{Qualitative results of ideal prompts, after adversarial attack, and after defense. Our method effectively restores segmentation quality under prompt degradation. }
	\label{result_ablation}
\end{figure}

\paragraph{Defense against adversarial attacks} We first evaluate the robustness of the defense agent against adversarial degradation. Specifically, we apply the trained attack agent to disrupt the ideal prompt configuration, followed by the defense agent to counteract the degradation. The final segmentation masks are produced by SAM based on the refined prompt set. As shown in the first three rows of Table~\ref{tab:attack_defense}, the attacker significantly reduces segmentation quality, indicating its ability to identify prompts that most adversely affect SAM’s output. In contrast, the defender effectively restores performance by removing these adversarial prompts, demonstrating its capacity to preserve prompts that are most beneficial for segmentation.

Qualitative results in Figure~\ref{result_ablation} further illustrate this process: segmentation accuracy clearly drops after the attack agent disrupts the ideal prompts, while it improves noticeably after the defense agent refines them. These results collectively validate the effectiveness of both agents.

\paragraph{Optimization of initial prompts} In real-world scenarios, initial prompt sets are often noisy or suboptimal. To simulate this, we generate initial prompts using a training-free feature matching strategy, and subsequently refine them using the trained defense agent. As shown in the last two rows of Table~\ref{tab:attack_defense}, SAM’s segmentation performance with raw prompts is limited. However, after refinement by PPD, segmentation quality improves consistently across all datasets, demonstrating the defender’s ability to enhance arbitrary prompt configurations in a plug-and-play fashion.

\begin{figure}[htp]
	\centering
	\caption{Qualitative results comparing initial prompts from the reference image and those optimized by PPD, which improves segmentation by removing disruptive prompts.}
	\includegraphics[width=0.93\linewidth]{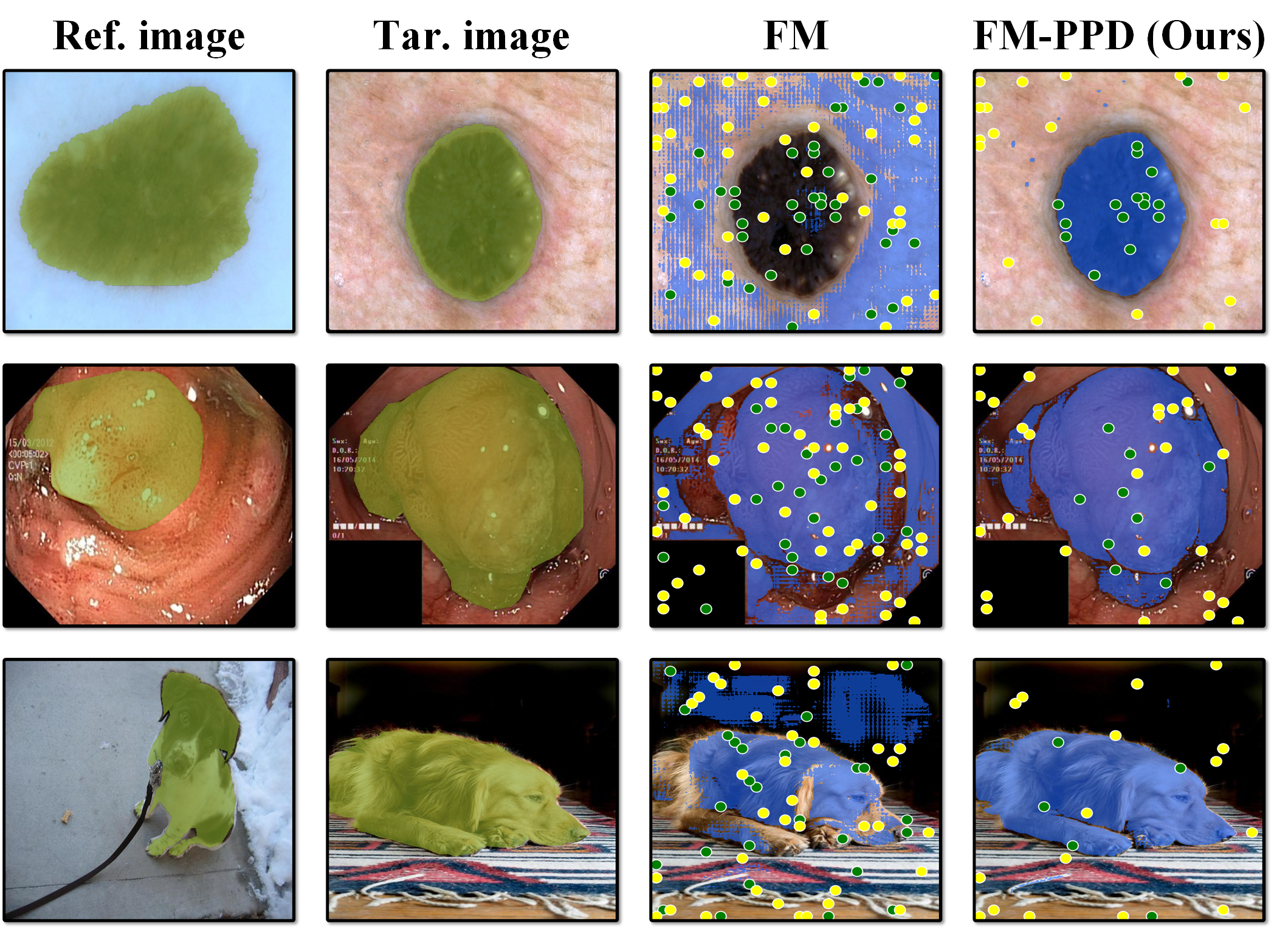} 
	\label{result_op}
\end{figure}

Qualitative results in Figure~\ref{result_op} further support this finding: across all three datasets, prompts generated from a single reference image via feature matching tend to include excessive or erroneous points, leading to poor segmentation. In contrast, PPD effectively improves prompt quality under a training-free paradigm by leveraging implicit structural cues in both physical and feature spaces, resulting in better segmentation outcomes.

\begin{table*}[t]
	
	\centering
		\caption{One-shot SAM-based segmentation performance on natural and medical datasets. \textbf{Bold} and \underline{underlined} values indicate the best and second-best results, respectively.}
	\begin{tabular}{lcccccc}
		\toprule
		\multirow{2}{*}{Methods} 
		& \multicolumn{2}{c}{PASCAL VOC}  
		& \multicolumn{2}{c}{ISIC}                                                     
		& \multicolumn{2}{c}{Kvasir}                                                                                                \\
		\cmidrule(lr){2-3} \cmidrule(lr){4-5} \cmidrule(lr){6-7} 
		& \multicolumn{1}{c}{mDSC (\%$\uparrow$)} 
		& \multicolumn{1}{c}{mIoU (\%$\uparrow$)} 
		& \multicolumn{1}{c}{mDSC (\%$\uparrow$)} 
		& \multicolumn{1}{c}{mIoU (\%$\uparrow$)} 
		& \multicolumn{1}{c}{mDSC (\%$\uparrow$)} 
		& \multicolumn{1}{c}{mIoU (\%$\uparrow$)} 
		\\ 
		\midrule
		
		PerSAM \cite{zhang2024personalize}   & 55.8 & 49.8 & 47.4 & 36.6 & 29.4 & 19.8  \\
		PerSAM-F   \cite{zhang2024personalize} & 50.0 & 44.0 & 59.6 & 50.4 & 25.0 & 18.3  \\
		Matcher \cite{liu2024matcher}        & \underline{68.9} & \textbf{60.4} & 69.6 & 60.9 & 35.0 & 25.3  \\
		VRP-SAM \cite{sun2024vrp}            & 49.1 & 39.9 & 4.6  & 2.6  & 0.0  & 0.0    \\
		GBMSeg \cite{liu2024featureprompting} & 61.3 & 52.7 & 53.8 & 40.0 & 33.9 & 22.9  \\
		FM-PPO \cite{liu2025plug}               & 62.1 & 53.7 & \underline{72.3} & \underline{62.4} & \underline{38.9} & \underline{29.5}  \\
		FM-PPD (Ours) &\textbf{69.1} &\underline{60.3}&\textbf{76.3}&\textbf{64.2}&\textbf{54.8}&\textbf{44.9} \\
		\bottomrule
	\end{tabular}

	\label{table_compare}
\end{table*}
\begin{figure*}[h]
	\centering
	\includegraphics[width=0.90\linewidth]{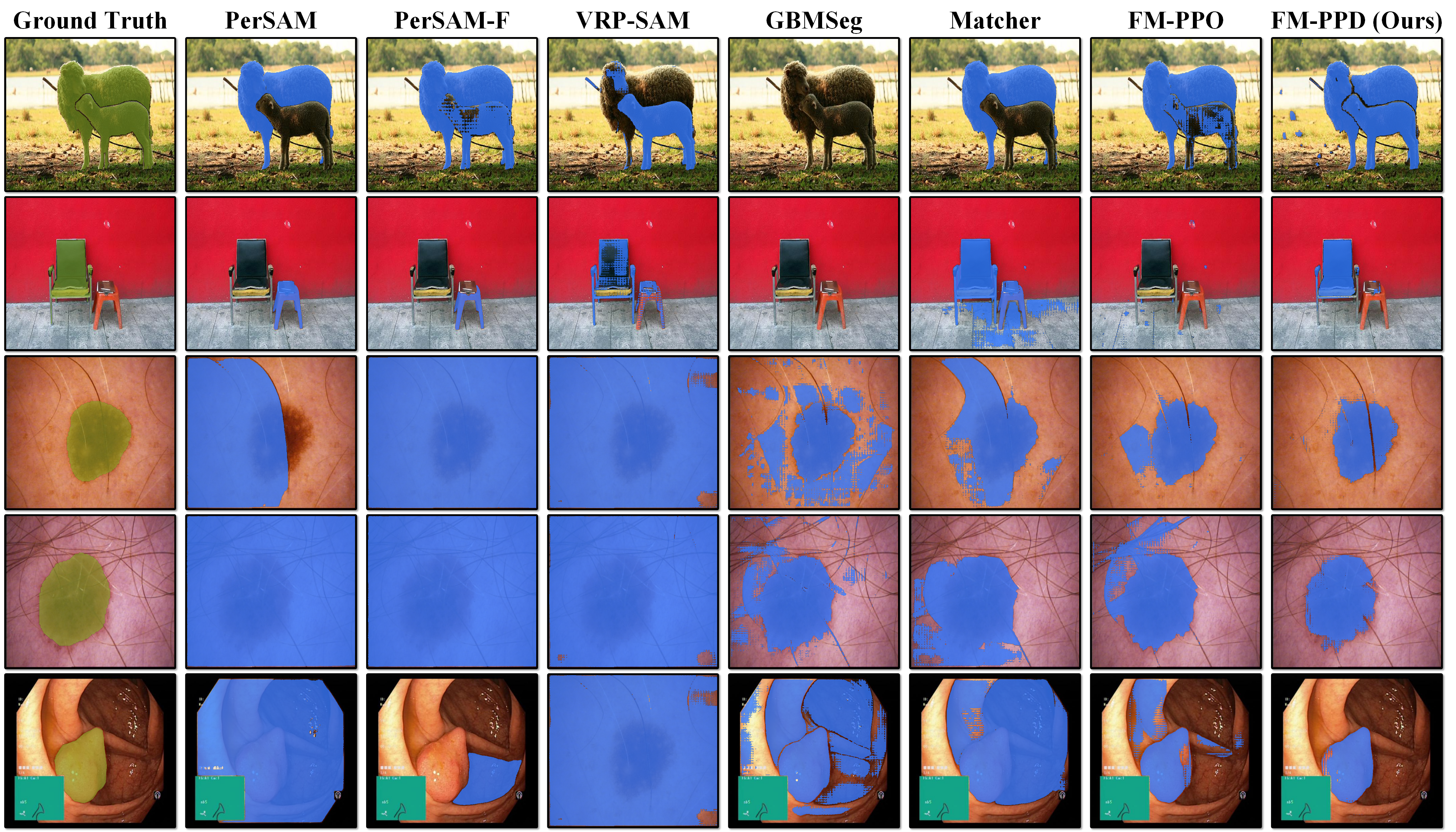} 
	\caption{Qualitative segmentation results of different one-shot SAM-based methods in natural and medical images.}
	\label{result_compare}
\end{figure*}

\subsection{Comparison with SAM-based One-shot Methods.}
To ensure a fair comparison with existing SAM-based segmentation methods, we adopt a unified one-shot setting across both natural and medical image datasets. In our framework, initial point prompts are automatically generated via feature matching between a reference image and each target image~\cite{liu2024matcher}, and the entire system is denoted as PPD-FM. Among all compared methods, only PerSAM-F performs task-specific fine-tuning on reference images within each dataset, while the others, including ours, operate in a training-free manner. As reported in Table~\ref{table_compare}, PPD-FM achieves consistently superior performance across domains without any task-specific retraining. The advantage is particularly evident on medical datasets with large domain shifts, where prompt encoders designed primarily for natural images—such as VRP-SAM—struggle to generalize.

In contrast, methods like PPO improve cross-domain performance through prompt optimization but remain sensitive to the quality of initial prompts due to limited diversity during training. PPD overcomes this limitation by introducing an adversarial dual-agent framework: the attack agent generates diverse and disruptive prompts to simulate poor prompts, while the defense agent learns to recover segmentation quality under such perturbations. This strategy encourages the defense agent to refine prompts in a feedback-driven manner, enhancing robustness and generalization without retraining.

Qualitative results in Figure~\ref{result_compare} further confirm the effectiveness of PPD-FM, which yields more accurate boundaries and suppresses irrelevant regions, even under suboptimal prompt conditions.




\section{Conclusion}

In this work, we present PPD, an adversarial reinforcement learning framework that automatically optimizes point prompts to enhance SAM’s segmentation performance. PPD models prompt interactions through a task-agnostic dual-space graph environment, constructed from both DINOv2-extracted feature distances and physical distances between points. It trains two adversarial agents: an attack agent that introduces disruptive prompts, and a defense agent that learns to suppress them. This general representation enables PPD to operate without task-specific supervision. Experiments on natural and medical image datasets demonstrate the effectiveness and generalization of our approach. Ablation studies show that the attack agent significantly degrades SAM’s performance by perturbing ideal prompts, while the defense agent reliably restores accuracy by filtering out harmful points. Moreover, the defense agent improves feature matching-based initial prompts and outperforms recent SAM-based segmentation methods across diverse domains. Overall, PPD offers a robust, task-agnostic, and plug-and-play solution for vision prompt optimization, providing a new perspective on adaptive enhancement of prompt-based segmentation models.

\paragraph{Limitations}
Despite achieving strong performance across diverse domains, PPD still relies on the availability of reasonably informative initial prompts. Although the defense agent can effectively refine noisy or redundant inputs, extremely poor initializations with limited spatial or semantic relevance may still affect segmentation quality. In future work, PPD can be combined with more advanced and adaptive prompt generation strategies, such as text-guided prompts or multimodal references, to further enhance its robustness in challenging cold-start scenarios.

\section*{Acknowledgments}
This should be a simple paragraph before the References to thank those individuals and institutions who have supported your work on this article.

\bibliographystyle{IEEEtran}
\bibliography{TMM}

\vfill

\end{document}